\documentclass[conference]{IEEEtran}
\IEEEoverridecommandlockouts
\usepackage{cite}
\usepackage{amsmath,amssymb,amsfonts}
\usepackage{algorithmic}
\usepackage{graphicx}
\usepackage{textcomp}
\usepackage{xcolor}
\usepackage{enumitem}
\usepackage{hyperref}
\hypersetup{
    colorlinks=true,
    linkcolor=blue,
    filecolor=magenta,
    urlcolor=cyan,
}
\def\BibTeX{{\rm B\kern-.05em{\sc i\kern-.025em b}\kern-.08em
    T\kern-.1667em\lower.7ex\hbox{E}\kern-.125emX}}
\begin{document}

\title{TraM : Enhancing User Sleep Prediction with Transformer-based Multivariate Time Series Modeling and Machine Learning Ensembles}

\author{\IEEEauthorblockN{Jinjae Kim}
\IEEEauthorblockA{\textit{Sch. of Business Administration} \\
\textit{Kookmin University}\\
Seoul, Korea \\
jinjae@kookmin.ac.kr}
\and
\IEEEauthorblockN{Minjeong Ma}
\IEEEauthorblockA{\textit{Sch. of Industrial and Management Engineering} \\
\textit{Korea University}\\
Seoul, Korea \\
minjeong\_ma@korea.ac.kr}
\and
\IEEEauthorblockN{Eunjee Choi}
\IEEEauthorblockA{\textit{Sch. of Electrical Engineering} \\
\textit{Korea University}\\
Seoul, Korea \\
eun09ji@korea.ac.kr}
\and
\IEEEauthorblockN{Keunhee Cho}
\IEEEauthorblockA{\textit{Civil and Environmental Engineering} \\
\textit{Korea Advanced Institute of Science and Technology}\\
Daejeon, Korea \\
keunhee0711@kaist.ac.kr}
\and
\IEEEauthorblockN{Changwoo Lee\IEEEauthorrefmark{1}\thanks{\IEEEauthorrefmark{1} Corresponding author: Changwoo Lee (leecw05@kookmin.ac.kr)}}
\IEEEauthorblockA{\textit{Sch. of Software} \\
\textit{Kookmin University}\\
Seoul, Korea \\
leecw05@kookmin.ac.kr}
}

\maketitle

\begin{abstract}
This paper presents a novel approach that leverages Transformer-based multivariate time series model and Machine Learning Ensembles to predict the quality of human sleep, emotional states, and stress levels. A formula to calculate the labels was developed, and the various models were applied to user data. Time Series Transformer was used for labels where time series characteristics are crucial, while Machine Learning Ensembles was employed for labels requiring comprehensive daily activity statistics. Time Series Transformer excels at capturing the characteristics of time series through pre-training, while Machine Learning Ensembles selects machine learning models that meet our categorization criteria. In experiments, the proposed model, TraM, scored 6.10 out of 10, demonstrating superior performance compared to other methodologies. The code and configuration for the TraM framework are available at: \href{https://github.com/jin-jae/ETRI-Paper-Contest}{https://github.com/jin-jae/ETRI-Paper-Contest}.
\end{abstract}

\begin{IEEEkeywords}
Time Series Transformer, Machine Learning Ensembles, Multivariate Time Series, MultiOutputClassfier
\end{IEEEkeywords}

\begin{table*}[htbp]
\caption{Explanation and Formulas of Labels}
\begin{center}
\begin{tabular}{|c|c|l|l|l|}
\hline
\textbf{Metric}&
    \textbf{Table}&
    \textbf{Explanation}&
    \textbf{Label Formula} \\ \hline
\textbf{Q1} &
  user\_survey &
  \begin{tabular}{@{}l@{}}
  Subjective sleep satisfaction assessment upon waking \\ 1 (Not at all) - 5 (Fully)
  \end{tabular} &
  sleep \\ \hline
\textbf{Q2} &
  user\_survey &
  \begin{tabular}{@{}l@{}}
  Pre-sleep emotional condition assessment \\ 1 (Very unpleasant) - 5 (Very pleasant)
  \end{tabular} &
  pmEmotion \\ \hline
\textbf{Q3} &
  user\_survey &
  \begin{tabular}{@{}l@{}}
  Pre-sleep stress level assessment \\ 1 (Very much) - 5 (Not at all)
  \end{tabular} &
  pmEmotion \\ \hline
\textbf{S1} &
  user\_sleep &
  Total sleep time &
  deepsleepduration + lightsleepduration + remsleepduration \\ \hline
\textbf{S2} &
  user\_sleep &
  Sleep efficiency &
  \begin{tabular}{@{}l@{}}
  (deepsleepduration + lightsleepduration + remsleepduration) \\ / (wakeupduration + deepsleepduration \\ + lightsleepduration +  remsleepduration) * 100
  \end{tabular} \\ \hline
\textbf{S3} &
  user\_sleep &
  Sleep onset latency (Time to sleep) &
  deepsleepduration + lightsleepduration + remsleepduration \\ \hline
\textbf{S4} &
  user\_sleep &
  Wake after sleep onset (Time to wake up) &
  wakeupduration - durationtosleep - durationtowakeup \\ \hline
\end{tabular}
\label{tab:formula}
\end{center}
\end{table*}

\section{Introduction}
The prevalence of sleep-related issues has led to a decline in sleep quality, increasing the risk of cardiovascular diseases and depression\cite{1_robillard2021profiles}. Poor sleep quality impacts physical and mental health, impairing cognitive function, weakening the immune system, and increasing illness vulnerability. Chronic sleep deprivation is linked to heightened stress, impaired judgment, and an increased risk of accidents, contributing to conditions such as hypertension, diabetes, obesity, and worsening mental health disorders.

Additionally, poor sleep quality is strongly associated with emotional regulation and stress. Chronic sleep deprivation leads to heightened stress levels, which further worsen sleep problems, creating a vicious cycle. Insufficient sleep disrupts the brain's ability to manage stress, resulting in increased irritability, anxiety, and mood swings. Studies show that sleep deprivation affects the emotional processing areas of the brain, such as the amygdala, leading to an overactive response to negative stimuli.\cite{16_cellini2020interplay}

The pandemic has led to lifestyle changes such as lockdowns, remote work, and increased screen time, significantly altering sleep patterns. Reduced physical exercise and irregular routines have disrupted circadian rhythms, making falling and staying asleep harder. Researchers have recognized the need to improve sleep quality by focusing on personalized solutions through technology and data analysis.

Understanding the relationship between sleep patterns and factors such as physical activity, stress, and environmental conditions can help develop targeted interventions. Improving sleep quality enhances productivity and contributes to long-term health benefits. Building a healthier society requires prioritizing sleep health in public health initiatives and individual wellness plans.

Chriskos et al.\cite{2_chriskos2021review} collected biometric signals, including skin temperature, heart rate, and respiratory rate to train a deep-learning model for predicting sleep quality. Ezati et al.\cite{3_ezati2020effect} showed that regular aerobic exercise improves sleep quality. Chang et al.\cite{4_chang2016association} found that individuals who perceive themselves as healthy or are satisfied with their physical activity levels tend to have better sleep quality. However, these models still require validation with diverse biometric data.

To address the issue, this paper proposes a novel approach combining a Time Series Transformer (TST)\cite{5_zerveas2021transformer} and Machine Learning Ensembles to predict sleep quality and related metrics. By integrating these advanced methodologies, our study aims to provide a robust framework for predicting and improving sleep quality. The combined strengths of TST and Machine Learning Ensembles offer an integrated solution for accurately predicting sleep quality and related life quality labels, contributing to improved health management and personalized intervention strategies.

The contributions of this paper are as follows:
\begin{itemize}
    \item Implemented methods for handling missing values, normalizing data, and ensuring temporal alignment in multivariate time series data.
    \item Refined a regression model to predict daily user responses based on historical time series data.
    \item Utilized ensemble method to improve the accuracy and robustness of life-quality predictions.
\end{itemize}

\section{Related Works}
Time series forecasting is essential in statistical analysis and machine learning. ARIMA\cite{6_shumway2017arima} combines autoregressive, moving average components, and differencing for stationarity, excelling in short-term predictions. Exponential Smoothing methods\cite{7_gardner2006exponential}, such as Holt’s Linear Trend Model and Holt-Winters Seasonal Model, apply exponentially decreasing weights to past data, effectively capturing trends and seasonality. Recurrent Neural Networks (RNNs)\cite{27_RNN_rumelhart1986learning} handle sequential data with hidden states incorporating previous time steps but struggle with long-term dependencies, addressed by Long Short-Term Memory networks\cite{29_lstm_hochreiter1997long} and Gated Recurrent Units\cite{30_gru_cho2014learning}. Convolutional Neural Networks\cite{28_cnn_lecun1989backpropagation}, adapted for time series with one-dimensional convolutions, often pair with RNNs to improve performance. Transformer models use self-attention mechanisms for capturing long-term dependencies, which are suitable for multivariate time series forecasting.

Machine Learning Ensembles enhances predictive modeling by combining algorithms for improved performance and robustness. Techniques include Bagging, Boosting, and Stacking. Bagging, like in Random Forests\cite{10_breiman2001random}, aggregates predictions from multiple decision trees\cite{14_quinlan1986induction} to reduce overfitting. Boosting, in algorithms such as Gradient Boosting Machines\cite{11_friedman2001greedy} and AdaBoost\cite{26_freund1997decision}, sequentially improves weak learners to form a strong model. Stacking trains a meta-model on predictions from multiple models. In time series forecasting and sleep quality prediction, Ensembles manages diverse data patterns and improves accuracy by capturing complex dependencies. Combining models like Support Vector Machines, Decision Trees, and K-Nearest Neighbors\cite{15_cover1967nearest} enhances predictions.

\section{Proposed Methods}
\begin{figure*}[htbp]
\includegraphics[width=\textwidth]{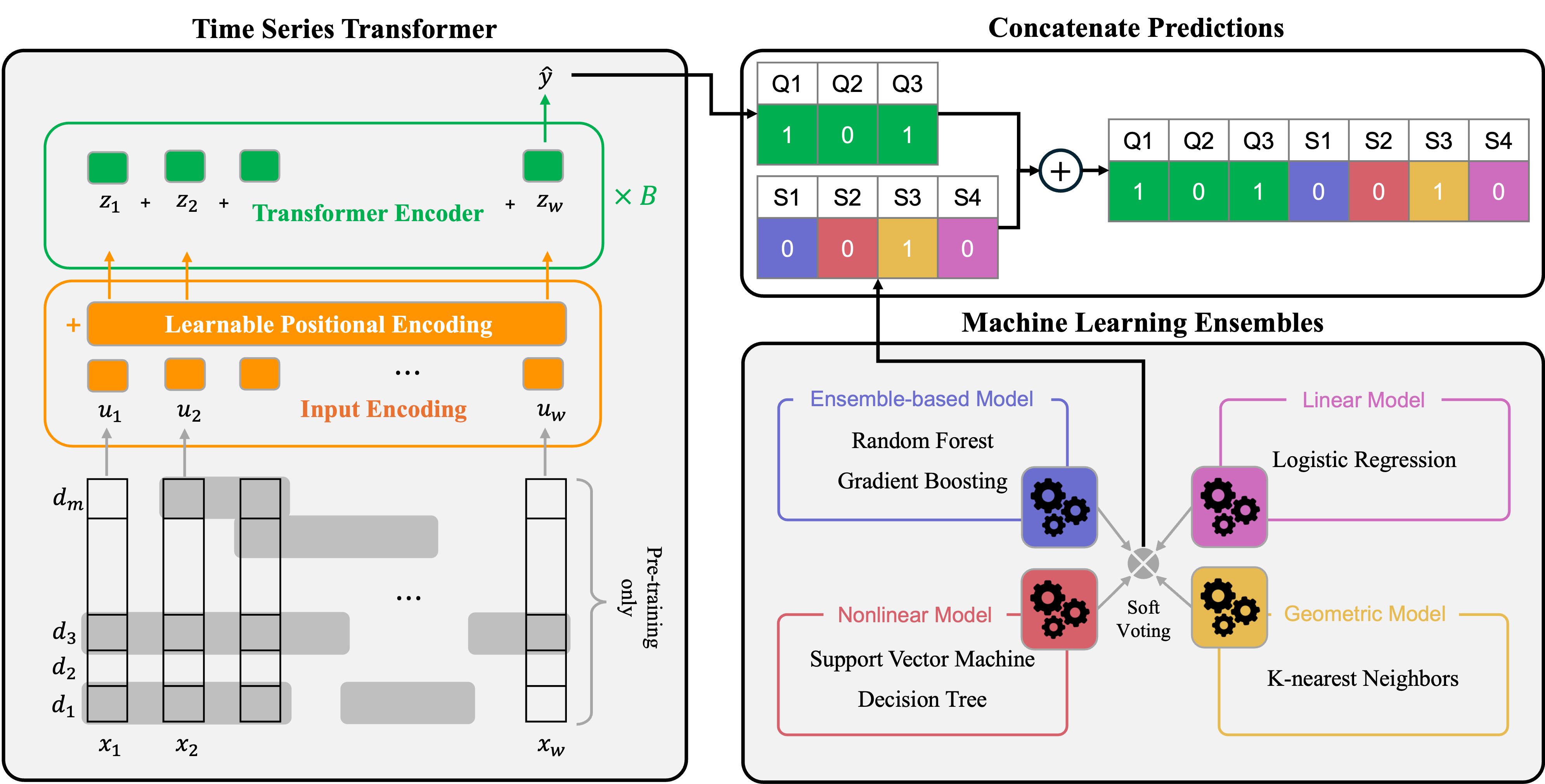}
\caption{Overview of our method. Our framework has two main components: a TST and Machine Learning Ensembles. On the left, TST utilizes a transformer encoder combined with learnable positional encoding and multiple input encoding blocks to process time series data. On the right, Machine Learning Ensembles shows the combination of various model types, integrated through soft voting to enhance prediction accuracy. We then concatenate Q1-Q3 labels from TST and S1-S4 labels from Machine Learning Ensembles.}
\label{fig:pipeline}
\end{figure*}

The objective of this competition\cite{17_oh2024human} was to predict both the Q1-Q3 labels, representing users' self-reported satisfaction via surveys and the S1-S4 labels, assessing sleep quality based on sensor data collected while sleeping. Before modeling, we examined the characteristics of each of the 7 labels. Using descriptions of each metric \cite{17_oh2024human} and the NSF sleep health guidelines \cite{18_ohayon2017national}\cite{19_hirshkowitz2015national}, we created formulas capable of predicting each label. For the Q1-Q3 labels, users' response on a particular day was assigned a 1 if it exceeded the average daily response of that user and a 0 if it was less. For the S1-S4 labels, a response is assigned a 1 if it meets the established threshold and a 0 if it does not. The detailed explanation and formula of labels can be found in TABLE ~\ref{tab:formula}.

The Q1, Q2, and Q3 labels represent ‘satisfaction with sleep,’ ‘emotional state before sleep,’ and ‘stress level before sleep’ respectively. These labels indicate an individual’s state which varies over time. It is crucial to consider the physical movements and changes resulting from daily activities. Capturing the emotional or stress levels derived from daily activity records in the training data is also essential. The time series data from daytime activities, which show clearer patterns and contents compared to nighttime, can be used to predict user responses based on sensor activity records. Given that specific users and dates as conditions, it is vital to reflect the time series characteristics particular to each user. The formula for the Q labels can be found in \eqref{eq_1}.

\begin{equation}
    Q_{i, d} =\begin{cases} 1 & \text{if } r_{i, d} > \mu_i \\0 & \text{if } r_{i, d} \leq \mu_i \end{cases}\label{eq_1}
\end{equation}

where $r_{i,d}$ denotes the survey value reported by a specific user $i$ on a specific day $d$, $\mu_i$ represents the daily average response of user $i$. Consequently, $Q_{i,d}$ is defined as the binary value assigned based on the conditions $i$ and $d$.

In contrast, the S1, S2, S3, and S4 labels represent specific, measurable sleep activities such as ‘total sleep time,’ ‘sleep efficiency,’ ‘time to fall asleep,’ and ‘time to wake up.’ These labels encompass numerous detailed aspects of sleep, such as its depth and continuity, which are not entirely recordable by smartphone or watch sensors in the dataset. The combinations of these columns are evaluated by comparing them to specific benchmarks. Given the granularity of the sensor data and the measured data, it is feasible to model the problem by classifying whether the conditions or states of a day meet these benchmarks. The formula for the S labels can be found in \eqref{eq_2}.

\begin{equation}
S = \begin{cases} 1 & \text{if } 7*60*60 < t < 9*60*60 \text{ (for S1)} \\1 & \text{if } p > 0.85*100 \text{ (for S2)} \\1 & \text{if } t < 30 *60 \text{ (for S3)} \\1 & \text{if } t < 20*60 \text{ (for S4)} \\0 & \text{otherwise}\end{cases}\label{eq_2}
\end{equation}

where $t$ represents the measurement time in seconds, $p$ represents the percentage, and $S$ denotes the binary value assigned to each user based on specific criteria.

\subsection{Time Series Transformer}
\subsubsection{Preprocessing}\label{tst_preprocessing}
Based on the Transformer Encoder, the TST model was used to predict the Q1-Q3 columns. This model is capable of performing both classification and regression tasks. Since the training and validation data had significant differences in their collection period and content, the data preprocessing step involved extracting common columns from both sets to create a model with high interpretability.

\begin{table}[htbp]
\caption{Preprocessed Features for TST}
\begin{center}
\begin{tabular}{|c|l|l|c|}
\hline
\begin{tabular}{@{}l@{}}
\textbf{Variable} \\ 
\textbf{Number}
\end{tabular} &
    \begin{tabular}{@{}l@{}}
    \textbf{Train} \\ 
    \textbf{Features}
    \end{tabular} &
    \begin{tabular}{@{}l@{}}
    \textbf{Validation} \\ 
    \textbf{Features}
    \end{tabular} &
    \begin{tabular}{@{}l@{}}
    \textbf{Extracted} \\ 
    \textbf{Methods}
    \end{tabular} \\ \hline
\textbf{1-12} &
    mAcc:x/y/z &
    m\_acc:x/y/z &
    \begin{tabular}{@{}l@{}}
    mean, std, \\ 
    min, max
    \end{tabular} \\ \hline
\textbf{13-20} &
    mGps:lat/lon &
    \begin{tabular}{@{}l@{}}
    m\_gps: \\
    latitude/longitude
    \end{tabular} &
    \begin{tabular}{@{}l@{}}
    mean, std, \\ 
    min, max
    \end{tabular} \\ \hline
\textbf{21-24} &
    e4Hr &
    w\_heart\_rate &
    \begin{tabular}{@{}l@{}}
    mean, std, \\ 
    min, max
    \end{tabular} \\ \hline
\textbf{25-33} &
    \begin{tabular}{@{}l@{}}
    \{timestamp\}\_label: \\
    activity
    \end{tabular} &
    m\_activity &
    \begin{tabular}{@{}l@{}}
    one hot \\ 
    labeled
    \end{tabular} \\ \hline
\end{tabular}
\label{tab:preprocess_tst}
\end{center}
\end{table}

To align the varied collection frequencies of each variable, we performed preprocessing to standardize the data collection periods. First, the collected data was resampled to a one-second interval, simply filling in any gaps with the preceding values to ensure a uniform time unit across all data. Then, to manage concerns about model size, the data was resampled again into ten-minute intervals for training. Since resampling from one-second to ten-minute intervals might not capture all the characteristics of the original data, we extracted the mean, standard deviation, minimum, and maximum for continuous variables. Additionally, for the categorical variable ‘m\_activity’, we applied one-hot encoding to capture the range of activities within each ten-minute interval, representing them using the maximum values. Ultimately, we were able to extract 128 days’ worth of sleep data, formatted in a way that the model could interpret based on daily sleep cycles. Since the test data had the same variable names as the validation data, we applied the same preprocessing method for both.

\subsubsection{Time Series Transformer Model}
TST Model \cite{5_zerveas2021transformer} is a Transformer-based model designed for multivariate time series classification and regression. This model employs the encoder part of the Transformer architecture and utilizes several techniques to effectively handle classification and regression tasks for time series data. Firstly, it incorporates a new type of positional encoding, specially designed to emphasize the sequential nature of time series data and capture temporal context. This enables the model to learn patterns effectively such as the cycle of the time series data. Secondly, the model generates input vectors using 1D-convolutional layers. This approach allows for enhanced performance on long, low-dimensional data. These filters extract essential features and adjust the dimensionality of the input. Additionally, modifying the stride or dilation of the filters facilitates effective data processing with large periodicities. Thirdly, a maximum sequence length is defined to address the issue of varying input data lengths inherent in time series data, and a padding mask is applied to shorter samples to maintain consistent input sizes. Finally, instead of layer normalization, batch normalization is employed to mitigate the impact of outliers in time series data. This allowed us to address the issue of anomalies in time series data, which could not be resolved using Transformer models.

\paragraph{Pre-training}
Pre-training involved the auto-regressive denoising task of predicting masked values in the input data, where masking followed a geometric distribution rather than a random assignment. This approach allowed the model to better reflect the structural characteristics of the input data. Such pre-training processes enable models to better understand the internal structure and correlations within the input data, offering the advantage of enhanced performance in downstream tasks such as classification and regression. According to results presented in TST, models that underwent supervised learning without pre-training generally recorded higher performance across most datasets than pre-trained ones. Pre-training was conducted using the preprocessed data specified in \ref{tst_preprocessing}.
\paragraph{Fine-tuning}
Fine-tuning was conducted on a pre-trained model by generating a long vector that concatenated representation vectors for each time step. This vector was then passed to a linear output layer. The model’s loss was calculated based on mean squared error, and the training was configured as a regression task. We trained the model as a regression, despite the final output being a classification of 0 or 1, because the threshold for classifying between 0 and 1 varied based on the average for each user. Consequently, actual user responses on a 1-5 scale for columns Q1, Q2, and Q3 were required for the fine-tuning process. However, among the provided datasets, only the train dataset contained actual user responses, while the validation dataset only included labels indicating whether responses exceeded the average. Therefore, only the train dataset was utilized for fine-tuning.

\subsection{Machine Learning Ensembles}
\subsubsection{Preprocessing}
Various machine learning methodologies were employed to analyze the correlations among multiple features and identify the factors that most significantly impact sleep quality. To create a more generalized model, features applicable to both training and validation datasets were explored. Given the vast amount of sensor data available, each user was sampled minute-by-minute, and the average and variance values of daily accelerometer, heart rate, activity, and GPS data were calculated.

Feature engineering was attempted using libraries such as tsfresh \cite{24_christ2018time} and various preprocessing techniques. Due to the tendency to overfit the training data with too many features, we ultimately proceeded with modeling using only the following 10 features. These overlapping ten columns in the training and validation datasets were all used for model training, and the evaluation data was used directly for prediction.

\begin{table}[htbp]
\caption{Preprocessed Features for MLE}
\begin{center}
\begin{tabular}{|c|l|l|c|}
\hline
\begin{tabular}{@{}l@{}}
\textbf{Variable} \\ 
\textbf{Number}
\end{tabular} &
    \begin{tabular}{@{}l@{}}
    \textbf{Train} \\ 
    \textbf{Features}
    \end{tabular} &
    \begin{tabular}{@{}l@{}}
    \textbf{Validation} \\ 
    \textbf{Features}
    \end{tabular} &
    \begin{tabular}{@{}l@{}}
    \textbf{Extracted} \\ 
    \textbf{Methods}
    \end{tabular} \\ \hline
\textbf{1-6} &
    mAcc:x/y/z &
    m\_acc:x/y/z &
    mean, var \\ \hline
\textbf{7} &
    \begin{tabular}{@{}l@{}}
    \{timestamp\}\_label: \\
    activity
    \end{tabular} &
    m\_activity &
    mode \\ \hline
\textbf{8} &
    e4Hr &
    w\_heart\_rate &
    mean \\ \hline
\textbf{9-10} &
    mGps:lat/lon &
    \begin{tabular}{@{}l@{}}
    m\_gps: \\
    latitude/longitude
    \end{tabular} &
    mean \\ \hline
\end{tabular}
\label{tab:preprocess_mle}
\end{center}
\end{table}

\subsubsection{MultiOutputClassifier}
Our goal was to predict multiple labels from a single dataset simultaneously. The MultiOutputClassifier provided by Sklearn is a representative methodology capable of predicting several labels, each with a cardinality of two. This model predicts the characteristics of samples where the features are not mutually exclusive, assigning a binary output to each class for all samples. This approach allows for the handling of multiple classes simultaneously and can explain interrelated behaviors.

Predictions were performed using six different models: RandomForestClassifier\cite{10_breiman2001random}, GradientBoostingClassifier\cite{11_friedman2001greedy}, LogisticRegression\cite{12_cox1958regression}, SVC\cite{13_boser1992training}, DecisionTreeClassifier\cite{14_quinlan1986induction}, and KNeighborsClassifier\cite{15_cover1967nearest}. Each model interprets data and performs predictions differently. By combining various models, it is possible to leverage the strengths and mitigate the weaknesses of each. For example, Decision Trees are sensitive to variables with high feature importance but are prone to overfitting, whereas KNN effectively reflects local characteristics but is sensitive to noise.

The six models can be broadly categorized into four types: ensemble-based model, linear model, nonlinear model, and geometric model. The first type includes ensemble-based models such as Random Forest and Gradient Boosting, which combine multiple decision trees or sequentially enhance weak learners, using a strategy to improve prediction performance by combining several base models. The second type includes the linear model, Logistic Regression, which uses a probabilistic approach through a linear decision boundary to predict outcomes. The third type comprises nonlinear models such as the Support Vector Machine and Decision Tree, which utilize the kernel trick to find the optimal separating hyperplane in high-dimensional space or identify branching points that effectively separate data, thus adeptly handling nonlinear data patterns. The fourth type is the geometric-based model, K-nearest Neighbors. It classifies based on the nearest k data points around a given data point, making it a model whose complexity directly depends on the data.

These varied models were combined, and soft voting was used to average the probabilities of each classifier’s predictions to make the final decision. If one model has high confidence in a particular class but others do not, soft voting incorporates this confidence as a probability, enabling more accurate predictions. Consequently, this allowed for more general and robust performance by considering the probabilities predicted by each model.

\section{Experiments}
\subsection{Experiment Setup}
Both models were utilized for the experiments on Ubuntu 18.04 and Python version 3.8.19, and dependencies were installed accordingly for each model. The training was conducted using two RTX 2080-Ti GPUs. During the pre-training phase of TST, the RAdam optimizer was used with a learning rate of 0.001 for 2000 epochs. The learning rate was set to 0.1 for 400 epochs during the fine-tuning. The batch size for both training phases was set to 128, and the model dimension (d\_model) was also set to 128. The random seed was fixed to 42 throughout the training process for both the Machine Learning Ensembles and TST to ensure reproducibility.

\subsection{Dataset}
The train dataset from ETRI\cite{25_chung2022real} contains sleep records for 508 days from 22 users. It includes sensor data measured by the e4 sensor and mobile devices, such as accelerometers, blood volume pressure, GPS, and gyroscopes, along with labeled activity information regarding the user’s location and type of activity. The validation and test datasets, also provided by ETRI, consist of sleep data for 105 and 115 days, respectively, for each of the 4 users. These datasets include sensor data measured by Android OS watches and smartphones, including accelerometer, light sensor, heart rate, and GPS. While the train dataset was collected in 2020, the validation and test dataset was collected in 2023, differing in their collection times and many of the sensors used. Both the train and validation datasets were used to extract common sensor data, which helped develop a model with high interpretability.

\subsection{Results}

\begin{figure}[htbp]\centerline{\includegraphics[width=\columnwidth]{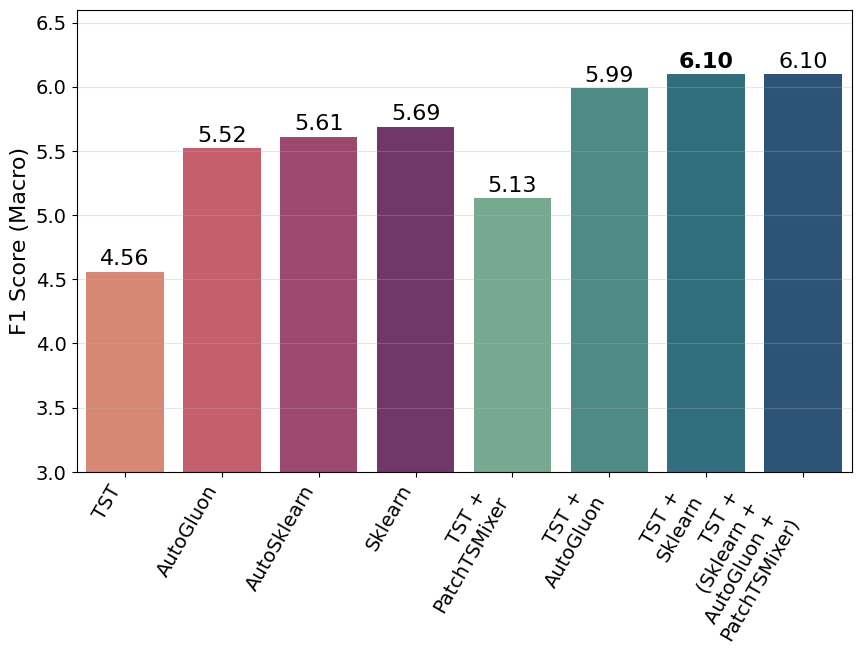}}
\caption{F1 Scores (Macro) on public test dataset}
\label{fig:experiment}
\end{figure}

\begin{table}[htbp]
\caption{F1 Score Comparison by Method}
\begin{center}
\begin{tabular}{|c|c|}
\hline
\textbf{Method} & \textbf{F1 Score (Macro)} \\
\hline
TST (Q1-S4) & 4.56 \\ \hline
AutoGluon (Q1-S4) & 5.52 \\ \hline
AutoSklearn (Q1-S4) & 5.61 \\ \hline
Sklearn (Q1-S4) & 5.69 \\ \hline
TST (Q1-Q3) + Sklearn (S1-S4) & \textbf{6.10} \\ \hline
\end{tabular}
\label{tab:f1_method}
\end{center}
\end{table}

Through the experimental results, we examined methodologies appropriate for the characteristics of each label. Using TST and Sklearn’s Machine Learning Ensembles to predict all labels resulted in low scores of 4.56 and 5.69, respectively. However, when using TST for Q1-Q3 and Sklearn for S1-S4, which are methodologies suitable for each label a higher performance of 6.10 was achieved. This confirms that Q1-Q3 benefit from methodologies where temporal characteristics are essential, and S1-S4 benefit from methodologies that independently classify numerical values by capturing all the daily features. Additionally, various AutoML libraries, such as AutoGluon and AutoSklearn, were used for prediction. AutoGluon \cite{20_erickson2020autogluon}, developed by Amazon, provides features to tune and ensemble deep learning and machine learning models and optimizes classification, regression, and forecasting tasks automatically, recording a performance of 5.52. AutoSklearn \cite{21_feurer2015efficient}, developed by the University of Freiburg in Germany, primarily selects machine learning models based on Sklearn and optimizes hyperparameters, recording a performance of 5.61. The results are summarized in TABLE~\ref{tab:f1_method}.

\begin{table}[htbp]
\caption{F1 Score Comparison by Sleep Prediction Model}
\begin{center}
\begin{tabular}{|c|c|}
\hline
\textbf{Method} & \textbf{F1 Score (Macro)} \\
\hline
TST + PatchTSMixer & 5.13 \\ \hline
TST + AutoGluon & 5.99 \\ \hline
TST + Sklearn & \textbf{6.10} \\ \hline
TST + (Sklearn+AutoGluon+PatchTSMixer) & 6.10 \\ \hline
\end{tabular}
\label{tab:f1_sleep}
\end{center}
\end{table}

Q1-Q3 were predicted using TST and various models were employed to predict S1-S4. The use of AutoGluon for label prediction yielded a very good result of 5.99, slightly improving on the performance of 5.52 obtained by using AutoGluon for all labels. PatchTSMixer \cite{22_ekambaram2023tsmixer} is a lightweight time series modeling technique based on the MLP-Mixer structure, which allows the model to be pre-trained and then used for downstream tasks such as time series prediction, classification, and regression. We used the classification model of PatchTSMixer, which showed a significantly lower performance of 5.13, reaffirming that basic machine learning methodologies are more effective for the S1-S4 labels. Finally, the results of the three models were ensembled using hard voting, achieving a score of 6.10. However, as this score did not significantly differ from the performance using only Sklearn, the model employing solely Sklearn was chosen for its cost-effectiveness. The results are summarized in TABLE~\ref{tab:f1_sleep}.

\begin{table}[htbp]
\caption{F1 Score Comparison by Data Preprocessing}
\begin{center}
\begin{tabular}{|c|c|}
\hline
\textbf{Method} & \textbf{F1 Score (Macro)} \\
\hline
TST + Sklearn (null processing) & 5.92 \\
\hline
TST + Sklearn (hyperparameter tuning) & 6.08 \\
\hline
TST + Sklearn & \textbf{6.10} \\
\hline
\end{tabular}
\label{tab:f1_preprocessing}
\end{center}
\end{table}

Predictions for Q1-Q3 were made using TST, followed by various modifications in the machine learning methodology S1-S4 predictions. The KNNImputer was utilized for handling missing values, and hyperparameters were optimized using grid search to enhance the performance of various classifiers. The data used in modeling, such as accelerometers, heart rate, activity, and GPS values, represent different types of data, which can have complex inherent relationships or vary in scale. Treating these heterogeneous data types similarly may have prevented finding appropriate neighbors. Replacing missing values by simply finding the nearest neighbors without considering temporal continuity led to a performance decrease to 5.92. Additionally, the performance slightly decreased to 6.08 after hyperparameter optimization, possibly due to overfitting caused by overly complex or deep model settings. Therefore, we reverted to using a basic model that replaces missing values with zeros and does not undergo hyperparameter tuning. The results are summarized in TABLE~\ref{tab:f1_preprocessing}.

\section{Conclusion}
In this paper, we introduce TST regression and Machine Learning Ensembles to predict user emotions, stress, and sleep quality. TST regression enhanced the accuracy of regression predictions for columns Q1, Q2, and Q3 through unsupervised pre-training. Meanwhile, Machine Learning Ensembles facilitated accurate classification predictions for columns S1, S2, S3, and S4. Using the public test dataset from the competition, we achieved an F1-Score(Macro) of 6.10 on a scale of 10.

\bibliographystyle{IEEEtran}
\bibliography{refs}

\begin{thebibliography}{10}
\providecommand{\url}[1]{#1}
\csname url@samestyle\endcsname
\providecommand{\newblock}{\relax}
\providecommand{\bibinfo}[2]{#2}
\providecommand{\BIBentrySTDinterwordspacing}{\spaceskip=0pt\relax}
\providecommand{\BIBentryALTinterwordstretchfactor}{4}
\providecommand{\BIBentryALTinterwordspacing}{\spaceskip=\fontdimen2\font plus
\BIBentryALTinterwordstretchfactor\fontdimen3\font minus \fontdimen4\font\relax}
\providecommand{\BIBforeignlanguage}[2]{{%
\expandafter\ifx\csname l@#1\endcsname\relax
\typeout{** WARNING: IEEEtran.bst: No hyphenation pattern has been}%
\typeout{** loaded for the language `#1'. Using the pattern for}%
\typeout{** the default language instead.}%
\else
\language=\csname l@#1\endcsname
\fi
#2}}
\providecommand{\BIBdecl}{\relax}
\BIBdecl

\bibitem{1_robillard2021profiles}
R.~Robillard, K.~Dion, M.-H. Pennestri, E.~Solomonova, E.~Lee, M.~Saad, A.~Murkar, R.~Godbout, J.~D. Edwards, L.~Quilty \emph{et~al.}, ``Profiles of sleep changes during the covid-19 pandemic: Demographic, behavioural and psychological factors,'' \emph{Journal of sleep research}, vol.~30, no.~1, p. e13231, 2021.

\bibitem{16_cellini2020interplay}
N.~Cellini and C.~Lombardo, ``The interplay between sleep and emotion: What role do cognitive processes play?'' p. 612498, 2020.

\bibitem{2_chriskos2021review}
P.~Chriskos, C.~A. Frantzidis, C.~M. Nday, P.~T. Gkivogkli, P.~D. Bamidis, and C.~Kourtidou-Papadeli, ``A review on current trends in automatic sleep staging through bio-signal recordings and future challenges,'' \emph{Sleep medicine reviews}, vol.~55, p. 101377, 2021.

\bibitem{3_ezati2020effect}
M.~Ezati, M.~Keshavarz, Z.~A. Barandouzi, and A.~Montazeri, ``The effect of regular aerobic exercise on sleep quality and fatigue among female student dormitory residents,'' \emph{BMC Sports Science, Medicine and Rehabilitation}, vol.~12, pp. 1--8, 2020.

\bibitem{4_chang2016association}
S.-P. Chang, K.-S. Shih, C.-P. Chi, C.-M. Chang, K.-L. Hwang, and Y.-H. Chen, ``Association between exercise participation and quality of sleep and life among university students in taiwan,'' \emph{Asia Pacific Journal of Public Health}, vol.~28, no.~4, pp. 356--367, 2016.

\bibitem{5_zerveas2021transformer}
G.~Zerveas, S.~Jayaraman, D.~Patel, A.~Bhamidipaty, and C.~Eickhoff, ``A transformer-based framework for multivariate time series representation learning,'' in \emph{Proceedings of the 27th ACM SIGKDD conference on knowledge discovery \& data mining}, 2021, pp. 2114--2124.

\bibitem{6_shumway2017arima}
R.~H. Shumway, D.~S. Stoffer, R.~H. Shumway, and D.~S. Stoffer, ``Arima models,'' \emph{Time series analysis and its applications: with R examples}, pp. 75--163, 2017.

\bibitem{7_gardner2006exponential}
E.~S. Gardner~Jr, ``Exponential smoothing: The state of the art—part ii,'' \emph{International journal of forecasting}, vol.~22, no.~4, pp. 637--666, 2006.

\bibitem{27_RNN_rumelhart1986learning}
D.~E. Rumelhart, G.~E. Hinton, and R.~J. Williams, ``Learning representations by back-propagating errors,'' \emph{nature}, vol. 323, no. 6088, pp. 533--536, 1986.

\bibitem{29_lstm_hochreiter1997long}
S.~Hochreiter and J.~Schmidhuber, ``Long short-term memory,'' \emph{Neural computation}, vol.~9, no.~8, pp. 1735--1780, 1997.

\bibitem{30_gru_cho2014learning}
K.~Cho, B.~Van~Merri{\"e}nboer, C.~Gulcehre, D.~Bahdanau, F.~Bougares, H.~Schwenk, and Y.~Bengio, ``Learning phrase representations using rnn encoder-decoder for statistical machine translation,'' \emph{arXiv preprint arXiv:1406.1078}, 2014.

\bibitem{28_cnn_lecun1989backpropagation}
Y.~LeCun, B.~Boser, J.~S. Denker, D.~Henderson, R.~E. Howard, W.~Hubbard, and L.~D. Jackel, ``Backpropagation applied to handwritten zip code recognition,'' \emph{Neural computation}, vol.~1, no.~4, pp. 541--551, 1989.

\bibitem{10_breiman2001random}
L.~Breiman, ``Random forests,'' \emph{Machine learning}, vol.~45, pp. 5--32, 2001.

\bibitem{14_quinlan1986induction}
J.~R. Quinlan, ``Induction of decision trees,'' \emph{Machine learning}, vol.~1, pp. 81--106, 1986.

\bibitem{11_friedman2001greedy}
J.~H. Friedman, ``Greedy function approximation: a gradient boosting machine,'' \emph{Annals of statistics}, pp. 1189--1232, 2001.

\bibitem{26_freund1997decision}
Y.~Freund and R.~E. Schapire, ``A decision-theoretic generalization of on-line learning and an application to boosting,'' \emph{Journal of computer and system sciences}, vol.~55, no.~1, pp. 119--139, 1997.

\bibitem{15_cover1967nearest}
T.~Cover and P.~Hart, ``Nearest neighbor pattern classification,'' \emph{IEEE transactions on information theory}, vol.~13, no.~1, pp. 21--27, 1967.

\bibitem{17_oh2024human}
S.~W. Oh, H.~Jeong, J.~M. Lim, S.~Chung, and K.~J. Noh, ``Human understanding ai paper challenge 2024--dataset design,'' \emph{arXiv preprint arXiv:2403.16509}, 2024.

\bibitem{18_ohayon2017national}
M.~Ohayon, E.~M. Wickwire, M.~Hirshkowitz, S.~M. Albert, A.~Avidan, F.~J. Daly, Y.~Dauvilliers, R.~Ferri, C.~Fung, D.~Gozal \emph{et~al.}, ``National sleep foundation's sleep quality recommendations: first report,'' \emph{Sleep health}, vol.~3, no.~1, pp. 6--19, 2017.

\bibitem{19_hirshkowitz2015national}
M.~Hirshkowitz, K.~Whiton, S.~Albert, C.~Alessi, O.~Bruni, L.~DonCarlos, N.~Hazen, J.~Herman, P.~Adams~Hillard, E.~Katz \emph{et~al.}, ``National sleep foundation’s updated sleep duration recommendations: final report. sleep health. 2015; 1 (4): 233--43,'' 2015.

\bibitem{24_christ2018time}
M.~Christ, N.~Braun, J.~Neuffer, and A.~W. Kempa-Liehr, ``Time series feature extraction on basis of scalable hypothesis tests (tsfresh--a python package),'' \emph{Neurocomputing}, vol. 307, pp. 72--77, 2018.

\bibitem{12_cox1958regression}
D.~R. Cox, ``The regression analysis of binary sequences,'' \emph{Journal of the Royal Statistical Society Series B: Statistical Methodology}, vol.~20, no.~2, pp. 215--232, 1958.

\bibitem{13_boser1992training}
B.~E. Boser, I.~M. Guyon, and V.~N. Vapnik, ``A training algorithm for optimal margin classifiers,'' in \emph{Proceedings of the fifth annual workshop on Computational learning theory}, 1992, pp. 144--152.

\bibitem{25_chung2022real}
S.~Chung, C.~Y. Jeong, J.~M. Lim, J.~Lim, K.~J. Noh, G.~Kim, and H.~Jeong, ``Real-world multimodal lifelog dataset for human behavior study,'' \emph{ETRI Journal}, vol.~44, no.~3, pp. 426--437, 2022.

\bibitem{20_erickson2020autogluon}
N.~Erickson, J.~Mueller, A.~Shirkov, H.~Zhang, P.~Larroy, M.~Li, and A.~Smola, ``Autogluon-tabular: Robust and accurate automl for structured data,'' \emph{arXiv preprint arXiv:2003.06505}, 2020.

\bibitem{21_feurer2015efficient}
M.~Feurer, A.~Klein, K.~Eggensperger, J.~Springenberg, M.~Blum, and F.~Hutter, ``Efficient and robust automated machine learning,'' \emph{Advances in neural information processing systems}, vol.~28, 2015.

\bibitem{22_ekambaram2023tsmixer}
V.~Ekambaram, A.~Jati, N.~Nguyen, P.~Sinthong, and J.~Kalagnanam, ``Tsmixer: Lightweight mlp-mixer model for multivariate time series forecasting,'' in \emph{Proceedings of the 29th ACM SIGKDD Conference on Knowledge Discovery and Data Mining}, 2023, pp. 459--469.

\end{thebibliography}

\vspace{12pt}
\end{document}